%% file: main.tex
\documentclass[runningheads]{llncs}

\usepackage{times}
\usepackage{epsfig}
\usepackage{graphicx}
\usepackage{adjustbox}
\usepackage{amsmath}
\usepackage{amssymb}
\usepackage{xcolor}

\newcommand{\ti}{\textit}

\usepackage{multirow}
\usepackage{epstopdf}
\usepackage{booktabs}

\begin{document}

\title{Combining Pyramid Pooling and Attention Mechanism for Pelvic MR Image Semantic Segmentaion}
\author{Ting-Ting Liang\inst{1}\and
Satoshi Tsutsui\inst{2} \and
Liangcai Gao\inst{1}\and
Jing-Jing Lu\inst{3}\and
Mengyan Sun\inst{3}}
\authorrunning{Liang et al.}
\institute{Peking University \and Indiana University Bloomington \and Peking Union Medical College Hospital}
\maketitle

\begin{abstract}
One of the time-consuming routine work for a radiologist is to discern anatomical structures from tomographic images. For assisting radiologists, this paper develops an automatic segmentation method for pelvic magnetic resonance (MR) images. The task has three major challenges 1)  A pelvic organ can have various sizes and shapes depending on the axial image, which requires local contexts to segment correctly. 2)  Different organs often have quite similar appearance in MR images, which requires global context to segment. 3) The number of available annotated images are very small to use the latest segmentation algorithms. To address the challenges, we propose a novel convolutional neural network called  Attention-Pyramid network (APNet) that effectively exploits both local and global contexts, in addition to a data-augmentation technique that is particularly effective for MR images. In order to evaluate our method, we construct fine-grained (50 pelvic organs) MR image segmentation dataset, and experimentally confirm the superior performance of our techniques over the state-of-the-art image segmentation methods.  

 \end{abstract}


\section{Introduction}
Medical doctors routinely identify the anatomical structure of human body from tomographic images, which is extremely time-consuming. In order to assist these doctors to understand tomographic images efficiently, it is one of the key research in medical imaging to develop a method to  automatically segment tomographic images into anatomical categories~\cite{1,2,3,4}. Among various tomography techniques, magnetic resonance (MR) imaging is often preferred for the purpose of radiotherapy planning due to the better soft-tissue contrast for organs involved in radiation therapy. 

In this paper, we are particularly interested in automatically segmenting pelvic MR images into 50 anatomical categories, which is much larger than previous work. The fine-grained segmentation results can greatly help radiologists to quickly identify pelvic structures, and be used for high-quality anatomical 3D reconstruction. In addition, the precise segmentation can help the doctors with follow-up diagnosis of relevant diseases such as sarcopenia. These are the primary motivations  that we want to develop a system that automatically segment pelvic structures. 

Our method is based on convolutional neural networks (CNNs), which is the backbone of state-of-the-art methods for image segmentation. However, segmenting pelvic MR images is more challenging than standard image segmentation due to its own characteristics. In fact,  it is often not an easy task, even for experienced doctors, to correctly segmenting pelvic MR images, especially when the images have unusual anatomical structures. The task requires a thorough comprehension of the pelvic anatomy, knowledge in the pelvic diseases that cause the unusual structure, and the ability to recognize patterns in the scanned images.  We collaborate with doctors in radiology department who actually segment pelvic MR images, and identify the three challenges for training CNNs. To address the challenges, we propose a novel CNN architecture called Attention-Pyramid network (APNet) and train it with a domain specific data augmentation. The challenges and our strategies are discussed in the following paragraphs. 

\begin{figure}[t]
  \centering
  \footnotesize
   \includegraphics[width=\textwidth]{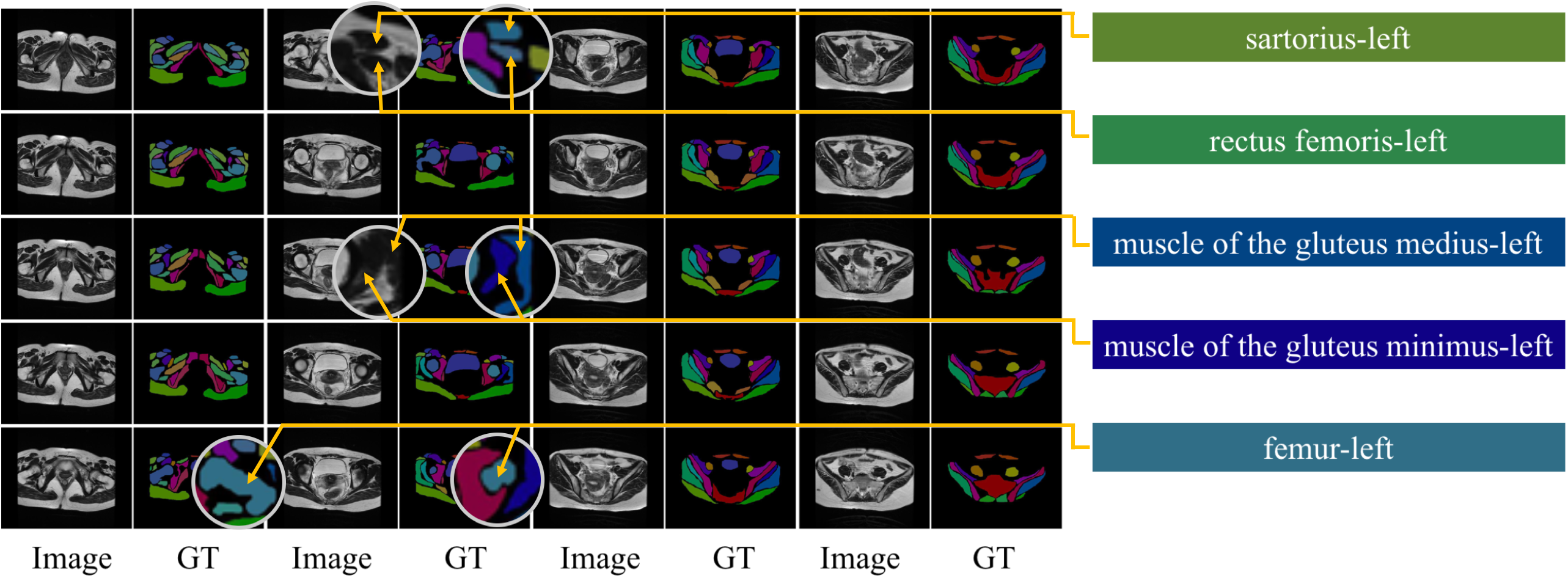}
  \caption{
  An exemplar series pelvic MR images belongs to one patient, each image presenting separate scanning sessions of different axial. Taking an image and its ground truth(GT) as a pair, the correct reading order is from top to bottom, from left to right.To stress our challenge, the magnified part of the image shows the characteristics of MR images.}
\label{fig:example}
\end{figure}

The first challenge is that a pelvic structure varies greatly in size and shape on different axial images, often with blurry boundaries caused by patient's belly movement when breathing.  For example, Fig.~\ref{fig:example} shows a series of representative MR images from a patient, where \textit{femur-left} has two completely different shapes in the second and fourth columns of the last row. Moreover, there is no clear boundary between \textit{muscle of the gluteus medius-left} and \textit{muscle of the gluteus minimus-left} in the middle row. For these cases, doctors often rely on multiple local contexts about the position of the organ (e.g. a particular organ should have two neighboring organs at bottom and right). In order for CNNs to effectively use these local contexts as doctors do, we adapt a layer that is particularly designed for multiple-level contexts aggregation, which is called a spatial pyramid pooling (see \S \ref{pyramid moule}).

The second challenge is that different pelvic organs often have very similar appearances in MR images. For example, in Fig.~\ref{fig:example}, \textit{sartorius-left} and \textit{rectus femoris-left} look very similar, the key to distinguish them is their positions: one on top and one on the bottom. For these cases, doctors usually depends on global contexts such as absolute positions of structures (e.g. doctors know that a particular organ should be always at the bottom-right of a pelvic image). To equip the similar ability for CNNs, we adopt an mechanism that is designed to gather global level context, which is called attention mechanism (see \S \ref{attention}). 


Third, the number of annotated images is limited. The annotation cost for the segmentation is much higher than other typical computer vision tasks such as image classification (i.e., tag annotation) or object detection (i.e., bounding box annotation). Furthermore,  unlike the natural images where we can use crowdsourcing for annotation, the medical task demands professional knowledge, which is not easily accessible.  Our dataset is composed of only 320 MR images from 14 patients. To address this problem, we apply elastic deformation to the annotated images, which is a type of data augmentation. This is an effective way especially for MR image segmentation, because image deformation often occurs in real MR images and thus realistic deformations can be simulated easily (see Fig.~\ref{fig:deform}). 

Our extensive experiments show that each technique (pyramid module, attention mechanism, and data augmentation) contributes to the better performance for pelvic MR image segmentation (see \S \ref{experiments}).

Overall, the main contributions of our work are: 
\begin{itemize}
\item We propose an automatic pelvic MR image segmentation method, which is the first one that completes pixel-level segmentation for a large number of structures (50 bones and muscles) on pelvic MR images.
\item We equip the network with a spatial pyramid pooling layer for aggregating the multiple level of local contexts. 
\item We build an attention-mechanism that effectively gathers global level of contexts. 
\item We adopt a data augmentation strategy with image deformation to increase realistic training images. 
\item We conduct extensive experiments to show the effectiveness of our proposed method. 
\end{itemize}

The rest of the paper is organized as follows. Section \ref{related_work} reviews relevant literature. Section \ref{methods} introduces the proposed model. Section \ref{experiments} presents the experimental results, and Section \ref{conclusion} concludes the paper. 

\input{related_work}

\input{model}

\input{experiments}

\section{Conclusion}\label{conclusion}
In this paper, we automatically segment MR pelvic images, with the goal to help medical professionals discern anatomical structures more efficiently. Our proposed methods address the three major challenges: high variation of organs' sizes and shapes often with the ambiguous boundaries, similar appearances of different organs, and small number of annotated MR images. To cope with these challenges, we propose Attention-Pyramid network and adopt a data augmentation strategy with image deformation. We experimentally demonstrate the effectiveness of our proposed methods over the baselines.

{\small
\bibliographystyle{splncs04}
\bibliography{reference}
}

\end{document}

%% file: related_work.tex

\section{Related work}\label{related_work}

\subsubsection{Pelvic segmentation}
\label{related_1}
Various methods for medical image segmentation have been developed over the past few years. Dowling et al.\cite{1} use an atlas-based prior method to 
detect the edges of hip-bone, prostate, bladder and rectum, and the Mean Dice Similarity Coefficient (DSC) of the four organs reached about 0.7. Ma et al.\cite{2} use a shape-guided Chan-Vese model, exploit the difference between pelvic organs' intensity distribution and simultaneously detect the edges of  bladder, vagina, rectum and levator ani muscle. In \cite{3}, MRI is used together with CT images to identify muscle structures, and the CycleGAN~\cite{27} is extended by adding a gradient consistency loss to improve the accuracy of the boundary. Kazemifar et al.\cite{4} use an encoder-decoder network called U-Net to segment male pelvic images, and it achieved 0.9 in Mean DSC. However, they segment sparsely distributed organs into four categories only. To the best of our knowledge, our work is the first one that classify each pixel in pelvic MR images into the fine-grained categories (50 bones and muscles) that are densely distributed, which is more challenging than previous work that typically segments sparsely distributed organs into a few categories. 
 
\subsubsection{Semantic segmentation in deep learning}\label{related_2}
Semantic segmentation is a task to classify every pixel in an image. Fully convolutional network~\cite{5} is the model that modifies image classification CNNs into semantic segmentation, and is a de facto backbone model for the state-of-the-art image segmentation. A problem in adopting CNN for segmentation is the existence of pooling layers. The pooling layer increases the receptive field by discarding the position information, but semantic segmentation requires pixel-wise classification, so the position information needs to be preserved. 

Researchers proposed two different forms of methods to address this problem.
The first is an encoder-decoder architecture such as U-Net~\cite{6} or SegNet~\cite{7}. The encoder uses the pooling layer to gradually reduce spatial dimensions of input data, and the decoder gradually recovers the details of target and the corresponding spatial dimensions through a network layer such as a deconvolution layer. It usually has a direct connection to pass information from encoder to decoder for better recovery of the position information. 

Another method is multi-scaling, which is the idea to use multiple sizes of input images (i.e., sharing network), convolutional filters (i.e., dilated convolution), or pooling layers (i.e., spatial pyramid pooling).  
The sharing network \cite{15,16} adjusts the size of input image to several proportions and passes them through a shared deep network. Then the final prediction result is from the  fusion of the resulting multi-scale features. 
Dilated convolution \cite{8,9,10} uses filters with multiple dilation (or atrous) factors, which can increase the receptive field without changing the size of feature map. 
Spatial pyramid pooling (SPP) \cite{11,12,13} divides input image into subregions, aggregates the characteristics of each sub-region, and finally concatenates features of all subregions to form a complete feature. This is an effective way to gather multiple levels of local contexts so we adapt it in our network. 

\subsubsection{Attention mechanism}\label{related_3}
Attention mechanism has been widely used in image processing. Xu et al.~\cite{21} introduce spatial visual attention mechanisms which extracts image features from the middle CNN layer. Jing et.al~\cite{18} propose an attention mechanism that learns the weights of both visual and semantic features, to define abnormal locations in medical images and generate relevant description sentences. While these methods apply attention mechanism in two-dimensional space or time dimension, we apply an attention mechanism for the scaling factors. Inspired from~\cite{18}, we propose an attention mechanism of joint learning, which combines predictions from multi-scale features when predicting the semantic label of a pixel. The final output of our model is generated by the maximum response of all scale predictions. We show that the proposed attention model effectively uses features at different locations and scales, which is crucial for identifying pelvic anatomical structures from global contexts.

%% file: model.tex

\section{Model}\label{methods}
\begin{figure*}[t]
	\centering
	\footnotesize
	\includegraphics[width=\textwidth]{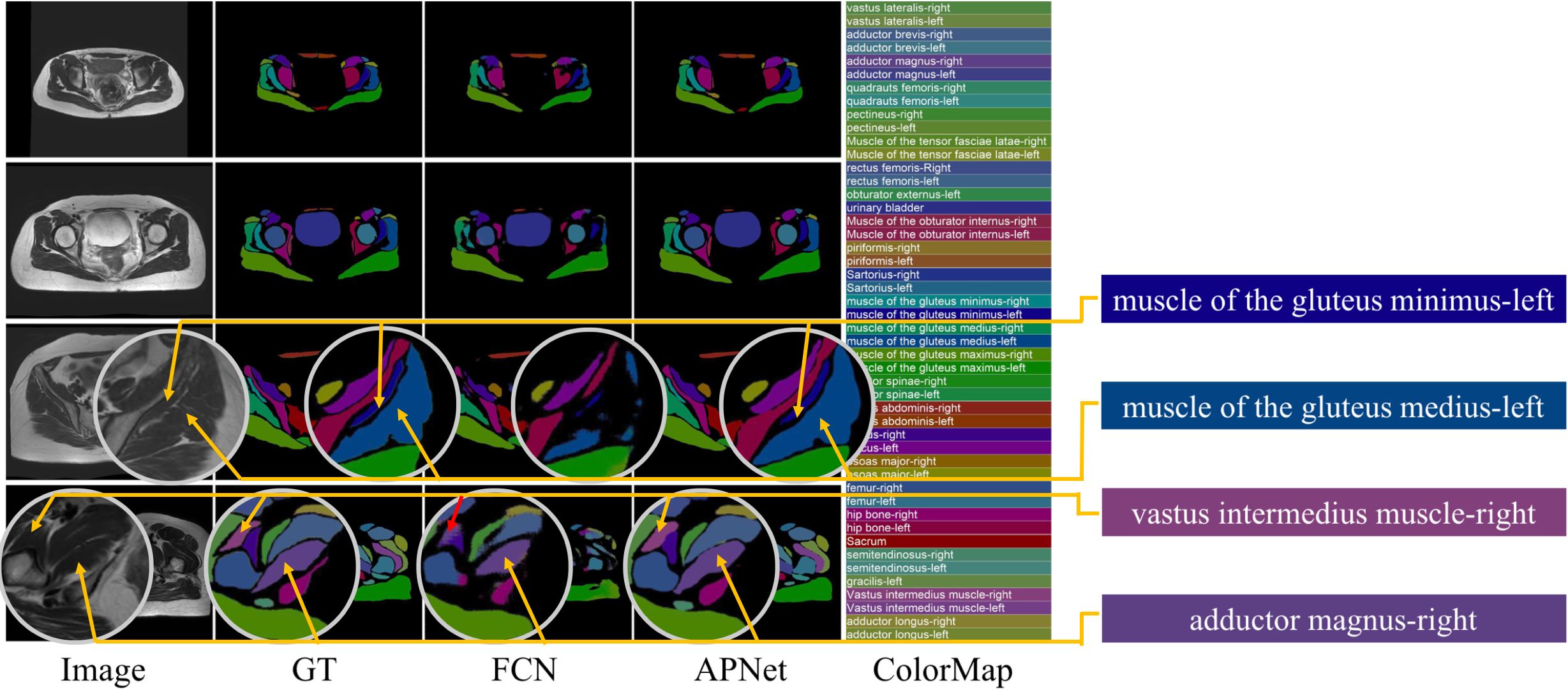}
	\caption{
Pelvic parsing issues we observe on our testset. FCN can only describe the structures roughly.
	}
	\label{fig:comparetofcn}
\end{figure*}

\begin{figure*}[t]
  \centering
  \footnotesize
  \includegraphics[width=\textwidth]{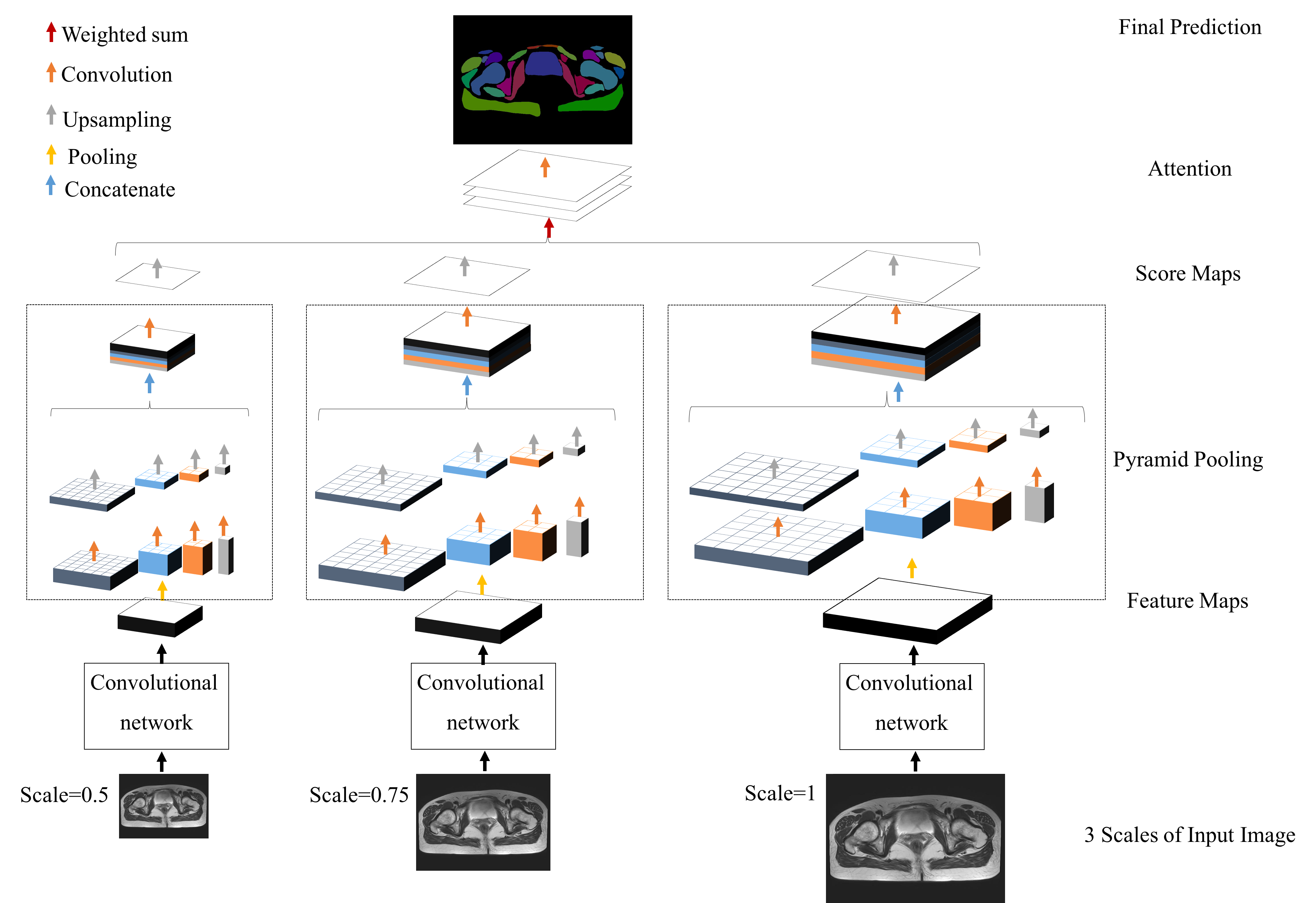}
  \caption{Illustration of the proposed APNet. First we resize input image into 3 scales with portion of ${\{1,0.75,0.5\}}$, and use CNN separately to get feature maps of 3 sizes from last convolutional layer.  Then a pyramid pooling layer, each for a feature map, forms pooled representation with bin sizes of ${1\times1, 2\times2, 3\times3, 6\times6}$ respectively. Followed by convolution, upsampling and concatenation, the four sub-region representations are concatenated with original feature map into a score map. Then score maps of 3 scales are upsampled to the maximum size, and the weighted sum of them gets the final score map. Finally, the representation is fed into a convolutional layer to get the final pixel-level prediction.}
 \label{fig:model}
\end{figure*}

This section describes our Attention-Pyramid Network (APNet) as illustrated in Fig.~\ref{fig:model}, which is designed to capture both local and global contexts. Our architecture engineering started from observing the segmentation results from FCN, which is a basic CNN for segmentation. In Fig~\ref{fig:comparetofcn}, we show samples from FCN and our APNet that we propose in this section. We can see that FCN fails to segment the boundary of \textit{muscle of the gluteus minimus-left} and \textit{muscle of the gluteus medius-left}. To segment this correctly, we need to care the local contexts of the two organs' boundaries. To equip this ability to CNN, we introduce spatial pyramid pooling that can capture multiple level of local contexts (See \S \ref{pyramid moule}). Moreover, we can also see FCN fails to distinguish \textit{vastus intermedius muscle-right} and \textit{adductor magnus-right}, which is due to the lack of the global context that \textit{vastus intermedius muscle-right} should be at lower place on this axial image. To recognize global context effectively, we introduce attention mechanism (see \S \ref{attention}). After technically describe the pyramid module and attention module, we finally describe the whole APNet architecture.


\subsection{Pyramid Pooling Layer}\label{pyramid moule}
Spatial Pyramid Pooling (SPP)~\cite{12} is to gather multiple levels of local contexts by pooling with multiple kernel sizes. Zhao et al.\cite{11} apply the SPP for FCN based segmentation to aggregate information across subregions of different scales (i.e, different level of local contexts). We adopt this SPP module for our pyramid pooling layer. The layer firstly separates feature map into different sub-regions and forms pooled representation for different positions. Assuming that a pyramid pooling layer has L levels, in each spatial bin, it pools the responses of each filter of input feature map under L level scales from course to fine. Assuming the input feature map has the size of ${n}{\times}{n}$, for one pyramid level of ${l}{\times}{l}$ bins, we implement this pooling level as a sliding window pooling, where the window kernel size =  $[1, n/l, n/l, 1]$, stride = $[1, n/l, n/l, 1]$, where ${[\cdot]}$ denotes ceiling operations. Each level reduces the dimension of feature map size to $1/l$ of the original one with level bin size of ${l}{\times}{l}$. Then we apply bilinear interpolation to upsample the low dimension features to the same size as the original feature map, and concatenate these features from multiple local contexts to form the final output of the pyramid pooling layer (See \S \ref{apnet}). 


\subsection{Attention}\label{attention}
Our attention mechanism is to capture the global contexts efficiently, and help the network to find the optimal weighting scheme for multiple input image sizes. We apply the attention mechanism for the output of pyramid pooling layer. Assuming we use $S$ scales (i.e., input image sizes), for ecah scale, the input image is resized and fed into a shared CNN that outputs a score map. The score maps from multiple scales are upsampled to the same size of the largest score map by bilinear interpolation. The final output is the weighted sum of score maps from all scales, where the weight reflects the importance of feature for each scale. The weighting scheme is initialized equally but, is updated by back-propagation in the training phase, so that it captures the global contexts effectively. 

Furthermore, to better merge discriminative features for the final convolutional layer output, we add extra supervision~\cite{23} for each scale in this attention mechanism. Lee et al.\cite{23} point out that distinguished classifiers trained with distinguished features demonstrate better performance in classification tasks. Particularly, the loss function for attention contains $1 + S$ cross entropy loss functions(one for final output and the other for each scale): 

\begin{equation}
F = \sum_{i=1}^{S}{\lambda_i}\cdot{F_i}
\end{equation}
\begin{equation}
Loss = min(H(F,gt)) + \sum_{i=1}^{S}min(H(F_i, gt_i))
\end{equation}
where $F$ denotes the final output, $gt$ denotes the original ground truth. $F_i$ is the score map of scale $s_i$ and $gt_i$ is the corresponding ground truth. The ground truths are downsampled accurately to the same size of corresponding outputs during training.  ${\lambda_i}$ is the weight of scale $s_i$. $H$ is cross entropy formula.

\subsection{Attention-Pyramid Network architecture }\label{apnet} With the attention mechanism on top of the pyramid pooling module, we propose our Attention-Pyramid network (APNet), as shown in Fig.~\ref{fig:model}. We use the three scales ${\{0.5, 0.75, 1\}}$, resize the input image for each scale, and feed the resized images to a shared CNN. We specifically use the ImageNet pre-trained ResNet \cite{22} with dilated network strategy \cite{8} to extract a feature map from \texttt{conv5} layer, which is  $1/8 $ of the input image in size. We feed the feature map into the pyramid pooling layer to gather multiple local contexts. The pooling window sizes are the whole, half,  $1/3 $, and  $1/6 $ of the feature map. Then we upsample the four pooled feature maps to the same size of original feature map, and concatenate them all. The pyramid pooling layer is followed by a convolutional layer to generate a score map for each scale . The weighted sum of three score maps will be the final segmentation results, where the weights are learned by the attention mechanism. 

APNet effectively exploits both local and global contexts for pixel-level pelvic segmentation. The pyramid pooling layer can collect local information and is more representative than the \textit{ global coordinator} \cite{13}. It learns local features and can adapt to the deformation of a structure on different axial images, but it sometimes confuses the categories (i.e., \textit{category confusion} \cite{11}) due to the lack of global contexts. This naturally calls for the attention mechanism that provides global contexts. The attention mechanism makes the model adaptively find the optimal weight for multiple scaled (or resized) images. Resizing does not change the relative size and position of organs, but smaller images helps CNNs to capture global contexts more easily than the high resolution images. Therefore, jointly training the attention mechanism and the spatial pooling layer is an effective way to gather both local contexts (by spatial pooling layer) and global contexts (by attention mechanism). 

We note that the global contexts include the absolute position of pelvic structures (left-right symmetry, up-and-down order) in the image. In other words, some organ categories are determined by the (global) position in the MR image. For example, a hint to recognize \textit{sartorius-right} is to check if it is on the right side of the image or not, which is exactly what radiologists does.

%% file: experiments.tex

\section{Experiments}\label{experiments}

\subsection{Datasets}
We prepare 320 MR images from 14 female patients, and professional doctors annotated them. The dataset covers image sizes of ${611}{\times}{610}, {641}{\times}{640}$, and $ {807}{\times}{582}$. Images belong to one patient is called a series. Each series has 24 or 20 images of same size presenting separate scanning sessions of different axes.
We use 240 MR images from 10 patients for training, 60 images from 3 patients for validation, and 20 MR images from a patient as test set.


\begin{figure}[t]
  \centering
  \footnotesize
  \includegraphics[width=0.7\textwidth]{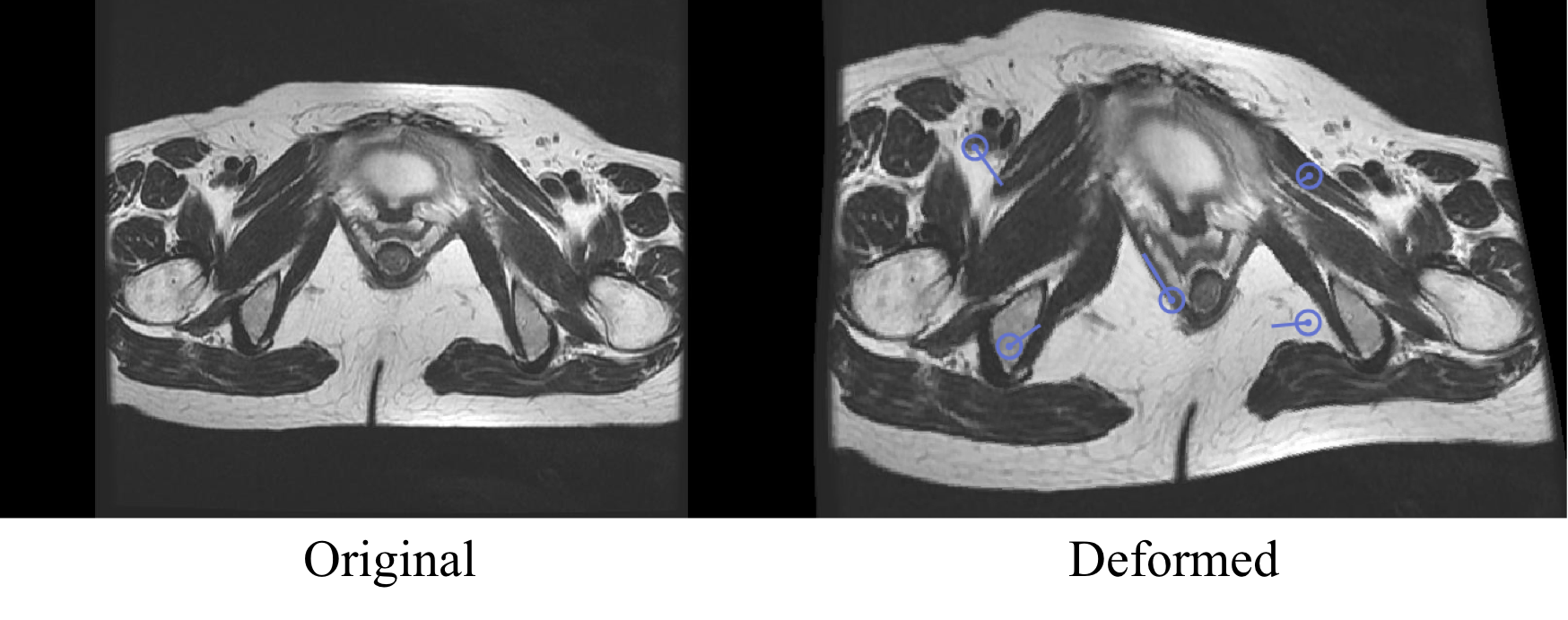}\vspace{-5mm}
  \caption{
 Example of image deformation. Blue circles are control points. 
}
\label{fig:deform}
\end{figure}

\subsection{Data Augmentation}\label{da}
For data augmentation, we originally tried random mirror, random resize, random rotation, and Gaussian Blur, which are effectively used in the state-of-the-art methods for natural scene parsing\cite{9,11}, but these conventional methods did not perform well. We call it common data augmentation (CDA) in the experiment section. In our work, we adopt image deformation using moving least squares\cite{24}. This is much more effective than the conventional methods, and can simulate one of the most common variation in MR images, because image deformation often occurs in real MR images. An example of data augmentation is shown in Fig.~\ref{fig:deform}. We perform random deformation multiple times on the dataset to get a training set with 30k MR images, a validation set with 2k MR images. We make sure that images from a patient are not in multiple sets.

\input{result_table.tex}

\subsection{Evaluation metrics}


Following \cite{5}, we use the pixel wise accuracy and region intersection over union(${IoU}$) between the segmentation results and ground truth to evaluate performance. Let ${TP_i}$(true positive) be the number of pixels of class $i$ predicted to belong to class $i$, ${FP_i}$(false positive) be the number of pixels of any other classes but $i$ predicted to be class $i$, ${FN_i}$(false negative) be the number of pixels of class $i$ predicted to belong to any other classes but $i$. Then we compute ${IoU_i}$ for class $i$ and ${mean\,IoU}$ for all classes:
\begin{itemize}
\item   \scalebox{0.8}{${IoU_i = TP_i / (TP_i + FP_i + FN_i)}$}
\item  \scalebox{0.8}{${Mean\,IoU = \sum_{i=1}^{n}(TP_i / (TP_i + FP_i + FN_i))}$}
\item \scalebox{0.8}{${Pixel\,Accuracy = \sum_{i=1}^{n}TP_i/(TP_i+FN_i)}$}
\end{itemize}


\subsection{Implementation details}
We use the Resnet-101 network pre-trained on Imagenet that is adapted for semantic segmentation as described in Section 3.2. To improve model speed, we reduce kernel size of resent101 conv5 from ${7\times7}$ to ${3\times3}$. Following \cite{9}, we implement dilated convolution with atrous sampling rate 12. For training, we adapt the  \ti{poly} learning rate policy~\cite{9} where current learning rate is multiplied by ${( 1-\frac{iter}{max_{iter}})^{power}}$. We set initial learning rate of ${2.5}{\times}{10^{-5}}$, power to $0.9$ respectively. With  iteration number of 110K in max, momentum and weight decay are set to $0.9$ and $0.0005$ respectively. We use Tensorflow~\cite{26} for implementation. 

\subsection{Baselines and Experimental Setup}

We compare APNet with three existing neural network architectures: FCN~\cite{5}, Deeplab-v2~\cite{9}, and PSPNet~\cite{11}. FCN is based on VGG architecture while others including ours are based on Dilated~\cite{10} Resnet-101\cite{16}. PSPNet is the state-of-the-art in natural image segmentation. Both APNet and PSPNet have spatial pyramid pooling~\cite{12} with the level of 4~\cite{11}. For APNet, we use two different levels of attention: $\{1, 0.75, 0.5\}$ and $\{1, 0.75\}$ where each number is a scaling factor for input image resizing. We intentionally use the factor more than 0.5 because it is known that scale portion less than 0.5 leads to unsatisfactory results~\cite{14}. The experiments with more levels of scaling factors are future work. 

All methods are trained with data augmentation strategies, as we describe in \S \ref{da}. To demonstrate the effectiveness of our data augmentation strategy, we also train PSPNet with common data augmentation (CDA) such as  random mirror, random resize, and random rotation, and call it \textit{PSPNet + CDA}. 

\subsection{Results and Discussions}

\begin{table}[t]
\centering
\small

\begin{tabular}{p{5cm}p{2cm}p{2cm}}
\hline

&mIoU(\%)	&Pixel Acc(\%)	\\
\hline\hline
FCN + DA &52.58 &72.03 \\
DeepLab-v2 + DA &76.70 &84.90 \\
PSPNet + DA	&76.08	&84.10	\\
PSPNet + CDA	&64.72	&74.39	\\
APNet (2 levels of attention) + DA	&79.38	&86.20	\\
APNet (3 levels of attention) + DA	&\textbf{80.27}	&\textbf{87.12}	\\
\hline

\end{tabular}
\caption{Test mIOU and mean pixel accuracy for each method. DA refers to deformation data augmentation based on image deformation. CDA refers to the common data augmentation.} \label{tab:com}
\end{table}


We show our experimental results on test set in Table~\ref{tab:main_paragraph}. APNet performs the best among FCN, DeepLab-v2, and PSPNet.  Of the two variants of APNet, 3 levels (${\{1, 0.75, 0.5\}}$) yields the best performance. With attention mechanism, our network has the highest score of $80.27 \%$ mIOU and $87.12 \%$ mean pixel accuracy. We can also see the benefit of our data augmentation (DA) strategy over the common data augmentation (CDA). When we train PSPNet with CDA, it only has $64.72 \%$ mIOU and  $74.39 \%$  mean pixel accuracy but with DA, it has $76.08 \%$ mIOU and  $84.10 \%$, which is more than $10 \%$ better results. 
\begin{figure}[t]
  \centering
  \footnotesize
  \includegraphics[width=\textwidth]{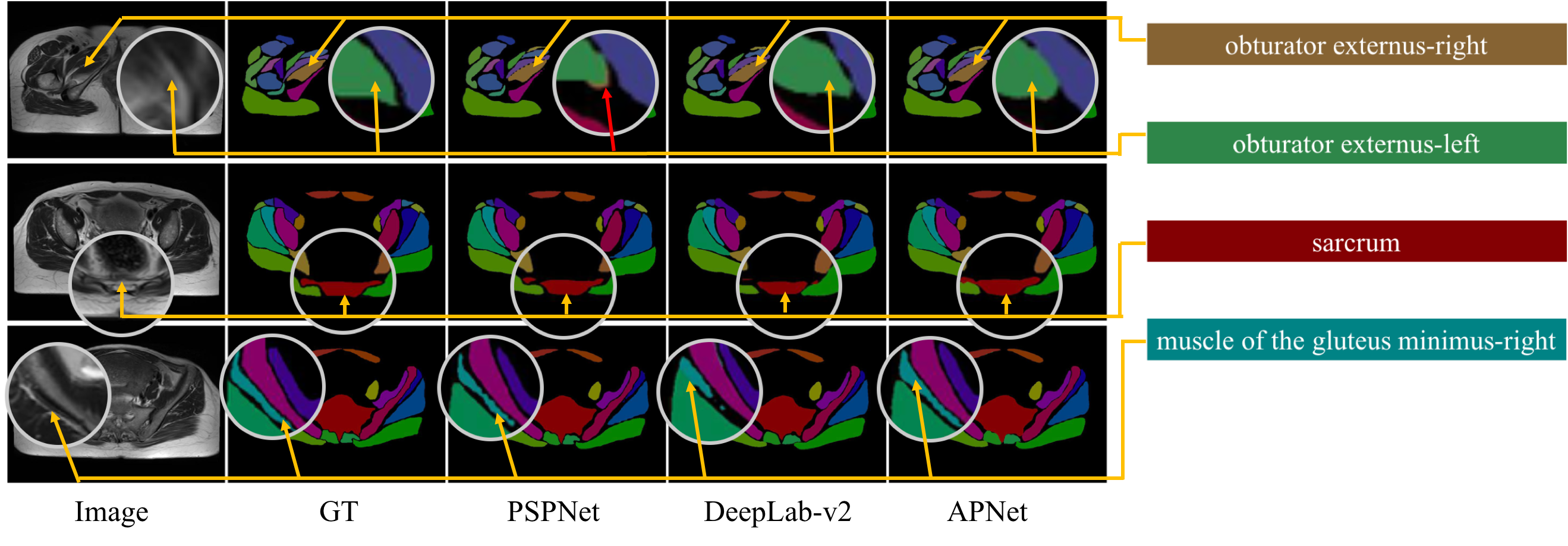}
  \caption{Improvements on test set. APNet captures more global context than PSPNet, and more local context than DeepLab-v2.
}
\label{fig:compares}
\end{figure}

We also show sample images from the test set in Fig~\ref{fig:compares}. APNet is designed to capture both local contexts and global contexts. The samples tell that DeepLab often fails to capture local  contexts, and PSPNet does not perform satisfactorily in capturing the global context, but APNet captures both. For example, in the first row, PSPNet misclassified a little part of \textit{obturator externus-left} to \textit{obturator externus-right}. APNet is able to fix the error due to the global context captured by attention mechanism. For another example, in the second row, we see that DeepLab-v2 failed to segment  \textit{sarcrum} precisely while APNet captured local context to describe its contour correctly. Similarly in the third row, DeepLab-v2 failed to capture the local context when segmenting the connected area of \textit{muscle of the gluteus minimus-right} while APNet can segment it precisely.

%% file: result_table.tex

\begin{table*}[htbp]
\centering
\begin{adjustbox}{max width=\textwidth}
\begin{tabular}{p{0.4cm}p{4.9cm}|p{1.3cm}p{1.6cm}p{1.1cm}|p{1.1cm}|p{2cm}p{2cm}}
\toprule[1pt]
No.	&Class name	&FCN+ DA	&DeepLab-v2 + DA	&{PSPNet + DA} 	&{PSPNet + CDA}	&{APNet     (2 levels of attention)  + DA}	&{APNet (3 levels of attention)  + DA}\\
\hline\hline										
								
1	&vastus lateralis-right	&56.73	&90.31	&91.44	&82.04	&91.12	&\textbf{92.68}	\\
\hline								
2	&vastus lateralis-left	&16.06	&74.75	&72.66	&57.27	&75.71	&\textbf{77.81}	\\
\hline								
3	&adductor brevis-right	&61.71	&\textbf{88.01}	&85.94	&70.07	&87.46	&87.85	\\
\hline								
4	&adductor brevis-left	&45.22	&82.35	&84.87	&74.76	&\textbf{88.51}	&87.26	\\
\hline								
5	&adductor magnus-right	&69.21	&87.72	&87.30	&73.70	&\textbf{89.48}	&\textbf{89.48}	\\
\hline								
6	&adductor magnus-left	&53.39	&88.55	&87.71	&79.94	&\textbf{91.23}	&90.99	\\
\hline								
7	&quadrauts femoris-right	&50.78	&74.77	&76.60	&60.17	&77.77	&\textbf{81.85}	\\
\hline								
8	&quadrauts femoris-left	&23.40	&69.65	&70.32	&52.37	&72.85	&\textbf{73.75}	\\
\hline								
9	&pectineus-right	&69.05	&80.29	&81.10	&62.90	&84.30	&\textbf{85.99}	\\
\hline								
10	&pectineus-left	&52.95	&77.62	&75.88	&62.13	&\textbf{79.98}	&79.28	\\
\hline								
11	&Muscle of the tensor fasciae latae-right	&70.74	&91.47	&91.18	&77.45	&\textbf{94.68}	&94.30	\\
\hline								
12	&Muscle of the tensor fasciae latae-left	&31.96	&72.28	&71.73	&56.84	&\textbf{76.05}	&74.08	\\
\hline								
13	&rectus femoris-Right	&58.38	&91.34	&92.02	&80.44	&\textbf{94.86}	&94.77	\\
\hline								
14	&rectus femoris-left	&35.62	&76.76	&77.33	&62.59	&79.25	&\textbf{80.18}	\\
\hline								
15	&obturator externus-right	&69.18	&88.16	&87.11	&71.61	&87.16	&\textbf{88.34}	\\
\hline								
16	&obturator externus-left	&65.64	&88.86	&87.71	&74.13	&89.68	&\textbf{90.56}	\\
\hline								
17	&urinary bladder	&92.15	&96.30	&95.26	&87.97	&96.36	&\textbf{96.52}	\\
\hline								
18	&Muscle of the obturator internus-right	&56.49	&74.51	&72.89	&53.95	&73.68	&\textbf{75.66}	\\
\hline								
19	&Muscle of the obturator internus-left	&51.02	&65.15	&61.87	&47.45	&65.70	&\textbf{66.62}	\\
\hline								
20	&piriformis-right	&71.52	&88.19	&87.30	&74.10	&89.01	&\textbf{89.10}	\\
\hline								
21	&piriformis-left	&72.00	&72.13	&67.69	&52.45	&71.63	&\textbf{74.27}	\\
\hline								
22	&Sartorius-right	&51.58	&81.98	&86.69	&73.75	&\textbf{89.59}	&89.40	\\
\hline								
23	&Sartorius-left	&27.23	&68.60	&69.33	&53.45	&\textbf{72.97}	&72.89	\\
\hline								
24	&muscle of the gluteus minimus-right	&55.44	&87.44	&89.62	&79.94	&91.97	&\textbf{92.82}	\\
\hline								
25	&muscle of the gluteus minimus-left	&29.01	&53.50	&53.19	&39.59	&58.28	&\textbf{58.3}	\\
\hline								
26	&muscle of the gluteus medius-right	&74.86	&93.20	&93.88	&85.31	&94.94	&\textbf{95.28}	\\
\hline								
27	&muscle of the gluteus medius-left	&48.44	&76.29	&74.55	&59.10	&77.59	&\textbf{77.84}	\\
\hline								
28	&muscle of the gluteus maximus-right	&86.61	&91.19	&91.25	&78.87	&91.76	&\textbf{93.65}	\\
\hline								
29	&muscle of the gluteus maximus-left	&81.25	&84.06	&82.12	&65.75	&84.02	&\textbf{84.70}	\\
\hline								
30	&erector spinae-right	&69.41	&77.64	&76.07	&56.49	&77.43	&\textbf{83.24}	\\
\hline								
31	&erector spinae-left	&62.81	&71.16	&69.88	&50.68	&72.35	&\textbf{73.26}	\\
\hline								
32	&rectus abdominis-right	&80.10	&81.52	&81.07	&68.65	&\textbf{88.45}	&86.69	\\
\hline								
33	&rectus abdominis-left	&77.56	&84.69	&81.63	&68.76	&\textbf{87.89}	&87.55	\\
\hline								
34	&iliacus-right	&67.93	&90.30	&92.20	&80.61	&\textbf{93.61}	&93.23	\\
\hline								
35	&iliacus-left	&48.71	&69.02	&64.92	&52.40	&69.34	&\textbf{71.25}	\\
\hline								
36	&psoas major-right	&66.03	&87.22	&84.47	&72.77	&88.90	&\textbf{88.9}	\\
\hline								
37	&psoas major-left	&48.53	&66.97	&64.74	&49.70	&69.72	&\textbf{69.72}	\\
\hline								
38	&femur-right	&80.97	&91.95	&91.70	&79.35	&92.51	&\textbf{93.99}	\\
\hline								
39	&femur-left	&72.57	&84.85	&83.72	&68.90	&85.10	&\textbf{86.35}	\\
\hline								
40	&hip bone-right	&73.26	&87.67	&86.92	&73.59	&88.67	&\textbf{89.70}	\\
\hline								
41	&hip bone-left	&62.82	&68.51	&64.85	&48.57	&68.52	&\textbf{70.13}	\\
\hline								
42	&Sacrum	&73.07	&79.82	&82.38	&72.16	&84.60	&\textbf{84.73}	\\
\hline								
43	&semitendinosus-right	&21.18	&65.73	&64.80	&49.30	&63.22	&\textbf{70.94}	\\
\hline								
44	&semitendinosus-left	&24.17	&56.17	&52.36	&44.85	&61.87	&\textbf{63.25}	\\
\hline								
45	&gracilis-right	&6.37	&46.88	&43.58	&29.04	&57.89	&\textbf{63.04}	\\
\hline								
46	&gracilis-left	&6.40	&43.25	&38.82	&30.77	&50.13	&\textbf{50.18}	\\
\hline								
47	&Vastus intermedius muscle-right	&9.82	&80.23	&79.06	&70.98	&85.42	&\textbf{86.46}	\\
\hline								
48	&Vastus intermedius muscle-left	&3.08	&63.02	&64.77	&55.44	&67.53	&\textbf{70.22}	\\
\hline								
49	&adductor longus-right	&50.87	&81.07	&83.32	&69.88	&\textbf{89.58}	&89.18	\\
\hline								
50	&adductor longus-left	&48.44	&78.53	&82.20	&72.67	&\textbf{87.90}	&87.19	\\
\hline\hline								
PixelAcc	&	&72.03	&84.90	&84.10	&74.39	&86.20	&\textbf{87.12}	\\
\hline								
mIoU	&	&52.58	&76.70	&76.08	&64.72	&79.38	&\textbf{80.27}	\\
\hline									

\toprule[1pt]
\end{tabular}
\end{adjustbox}
\caption{Per-class IoU(\%) on test set. \ti{CDA} refers to common data augmendation. \ti{DA} refers to the deformation data augmentation we performed, \ti{levels of attention} indicates how many different sizes we adjust for the input image.}
\label{tab:main_paragraph}
\end{table*}

%% file: main.bbl
\begin{thebibliography}{10}
\providecommand{\url}[1]{\texttt{#1}}
\providecommand{\urlprefix}{URL }
\providecommand{\doi}[1]{https://doi.org/#1}

\bibitem{26}
Abadi, M., Barham, P., Chen, J., Chen, Z., et.al: Tensorflow: A system for
  large-scale machine learning. 12th USENIX Symposium on Operating Systems
  Design and Implementation (OSDI)  (2016)

\bibitem{7}
Badrinarayanan, V., Kendall, A., Cipolla, R.: Segnet: A deep convolutional
  encoder-decoder architecture for image segmentation. arXiv:1511.00561  (2015)

\bibitem{9}
Chen, L.C., Papandreou, G., Kokkinos, I., Murphy, K., Yuille, A.L.: Deeplab:
  Semantic image segmentation with deep convolutional nets, atrous convolution,
  and fully connected crfs. TPAMI  (2016)

\bibitem{14}
Chen, L.C., Yang, Y., Wang, J., Xu, W., Yuille, A.L.: Attention to scale:
  Scale-aware semantic image segmentation. CVPR  (2016)

\bibitem{16}
Ciresan, D., Meier, U., Schmidhuber, J.: Multi-column deep neural networks for
  image classification. CVPR  (2012)

\bibitem{1}
Dowling, J.A., Lambert, J., Parker, J., Salvado, O., Fripp, J., Capp, A.,
  Wratten, C., Denham, J.W., Greer, P.B.: An atlas-based electron density
  mapping method for magnetic resonance imaging (mri)-alone treatment planning
  and adaptive mri-based prostate radiation therapy. International Journal of
  Radiation Oncology  (2011)

\bibitem{22}
He, K., Zhang, X., Ren, S., Sun, J.: Deep residual learning for image
  recognition. CVPR  (2016)

\bibitem{12}
He, K., Zhang, X., Ren, S., Sun, J.: Spatial pyramid pooling in deep
  convolutional networks for visual recognition. TPAMI  (2015)

\bibitem{3}
Hiasa, Y., Otake, Y., Takao, M., Matsuoka, T., azuma Takashima, Prince, J.L.,
  Sugano, N., Sato, Y.: Cross-modality image synthesis from unpaired data using
  cyclegan: Effects of gradient consistency loss and training data size. MICCAI
   (2018)

\bibitem{18}
Jing, B., Xie, P., Xing, E.: On the automatic generation of medical imaging
  reports. CVPR  (2018)

\bibitem{4}
Kazemifar, S., Balagopal, A., Nguyen, D., McGuire, S., Hannan, R., Jiang, S.,
  Owrangi, A.: Segmentation of the prostate and organs at risk in male pelvic
  ct images using deep learning. arXiv:1802.09587  (2018)

\bibitem{23}
Lee, C.Y., S.~Xie, P.G., Zhang, Z., Tu, Z.: Deeply-supervised nets. AISTATS
  (2015)

\bibitem{13}
Liu, W., Rabinovich, A., Berg, A.C.: Parsenet: Looking wider to see better.
  ICLR  (2016)

\bibitem{5}
Long, J., Shelhamer, E., Darrell, T.: Fully convolutional networks for semantic
  segmentation. CVPR  (2015)

\bibitem{2}
Ma, Z., Jorge, R.N.M., Mascarenhas, T., Tavares, J.M.R.S.: Segmentation of
  female pelvic cavity in axial t2-weighted mr images towards the 3d
  reconstruction. Int J Numer Method Biomed Eng.
  \textbf{28(6-7)}(10.1002/cnm.2463),  714--26 (2012)

\bibitem{15}
P.F.Felzenszwalb, R.B.Girshick, D.McAllester, D.Ra-manan.: Object detection
  with discriminatively trained part- based models. TPAMI  (2010)

\bibitem{6}
Ronneberger, O., Fischer, P., Brox, T.: U-net: Convolutional networks for
  biomedical image segmentation. MICCAI  (2015)

\bibitem{24}
Schaefer, S., McPhail, T., Warren, J.: image deformation using moving least
  squares. ACM Transactions on Graphics (TOG) - Proceedings of ACM SIGGRAPH
  (2006)

\bibitem{21}
Xu, K., Ba, J., Kiros, R., Cho, K., Courville, A., Salakhudi-nov, R., Zemel,
  R., Bengio, Y.: Show, attend and tell: Neural image caption generation with
  visual attention. ICML  (2015)

\bibitem{8}
Yu, F., Koltun., V.: Multi-scale context aggregation by dilated convolutions.
  arXiv:1511.07122  (2015)

\bibitem{10}
Yu, F., Koltun, V., Funkhouser, T.: Dilated residual networks. CVPR  (2017)

\bibitem{11}
Zhao, H., Shi, J., Qi, X., Wang, X., Jia, J.: Pyramid scene parsing network.
  CVPR  (2017)

\bibitem{27}
Zhu, J.Y., e.a.: Unpaired image-to-image translation using cycle-consistent
  adversarial networks. CVPR  (2017)

\end{thebibliography}
